\documentclass[conference]{IEEEtran}
\IEEEoverridecommandlockouts
\usepackage{multirow}
\usepackage{booktabs}
\usepackage{graphicx}
\usepackage{ulem}
\usepackage{soul}
\usepackage{xcolor}

\usepackage{cite}
\usepackage{amsmath,amssymb,amsfonts}
\usepackage{algorithmic}
\usepackage{graphicx}
\usepackage{textcomp}
\usepackage{xcolor}
\def\BibTeX{{\rm B\kern-.05em{\sc i\kern-.025em b}\kern-.08em
    T\kern-.1667em\lower.7ex\hbox{E}\kern-.125emX}}
\begin{document}

\title{Graph Inference Towards ICD Coding\\
%
}

\author{
\IEEEauthorblockN{1\textsuperscript{st} 
Xiaoxiao Deng}
\IEEEauthorblockA{\textit{College of Computing and Digital Media} \\
\textit{DePaul University}\\
Chicago, United States  \\
dengxiaoxiao1019@gmail.com}
}

\maketitle

\begin{abstract}
Automated ICD coding involves assigning standardized diagnostic codes to clinical narratives. The vast label space and extreme class imbalance continue to challenge precise prediction. To address these issues, LabGraph is introduced—a unified framework that reformulates ICD coding as a graph generation task. By combining adversarial domain adaptation, graph-based reinforcement learning, and perturbation regularization, LabGraph effectively enhances model robustness and generalization. In addition, a label graph discriminator dynamically evaluates each generated code, providing adaptive reward feedback during training Experiments on benchmark datasets demonstrate that LabGraph consistently outperforms previous approaches on micro-F1, micro-AUC, and P@K.
\end{abstract}


\section{Introduction}
\begin{figure}[b]
\centering
\includegraphics[width=1\columnwidth]{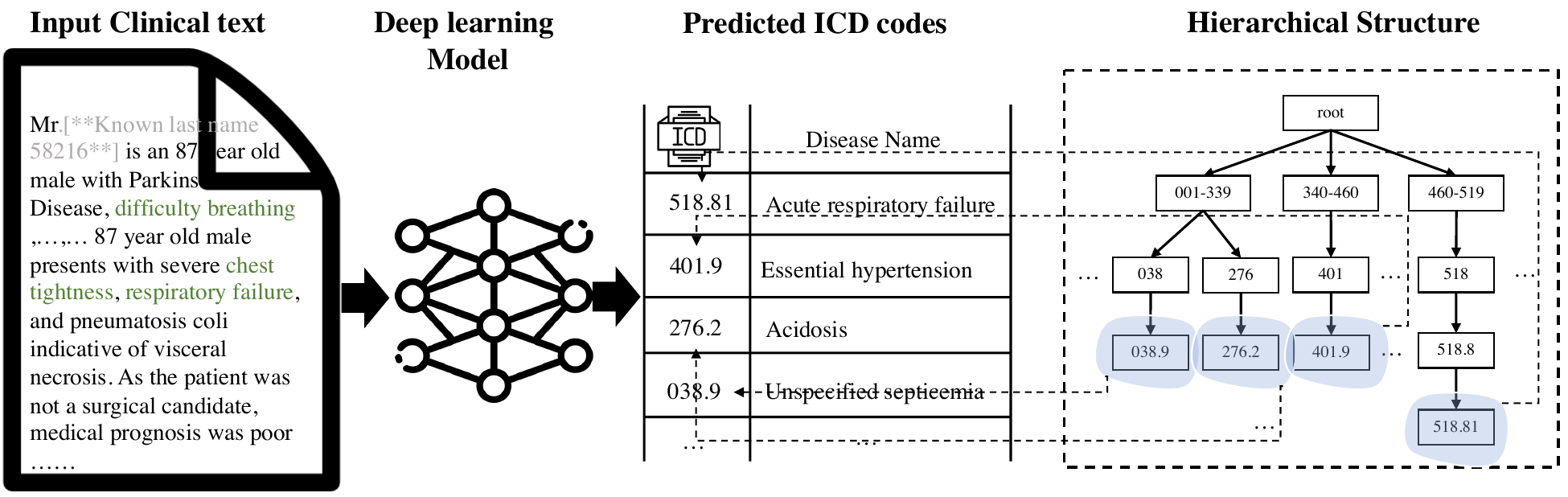} 
\caption{ICD-9 code hierarchy and example of automatic ICD coding, with clinical text as input and predicted codes as output.}
\label{fig: codes}
\end{figure}

Automated ICD coding has gained attention for its potential to reduce manual effort in clinical documentation and billing. Patient records contain diverse information, including admissions, physician notes, medical history, and lab tests \cite{nie2024towards}. As illustrated in Figure~\ref{fig: codes}, ICD codes are organized hierarchically, with sibling codes rarely co-occurring in a single record. This sparse, hierarchical structure poses challenges for models, which must capture both semantic relationships and code dependencies for accurate prediction.

Automated ICD coding is typically handled as a multi-label
prediction task using neural network techniques \cite{cao-etal-2020-hypercore}. Automated ICD coding faces several key challenges. First, the label distribution is extremely imbalanced: out of 9,219 possible codes, only 122 appear frequently within the top 50 categories, leaving the majority of codes sparsely represented in clinical records. Second, crucial connections between codes—like
sibling associations, mutually exclusive pairs, and hierarchical
parent-child links—are frequently overlooked or inadequately
modeled \cite{nelson2024icd}. For instance, ``783.1'' (Abnormal weight gain) and ``783.2'' (Abnormal weight loss) cannot logically co-occur, yet such constraints are rarely considered in existing systems. Third, relying on a single training strategy can limit model performance, particularly for rare disorders that appear infrequently in the data \cite{vu2020label}. 

To address challenges in automated ICD coding, we reformulate the task as labeled graph generation over the ICD code hierarchy. A global code graph is first constructed from training labels, and initial labels are generated from the root node based on clinical text input. Subsequent labels are predicted using neighboring nodes and embeddings, reducing the candidate space. We propose LabGraph, a multi-component framework with four training strategies, comprising six modules: Label Graph Generator (LGG), Label Graph Discriminator (LGD), Message Integration Module (MIM), MHR-CNN, Fat-RGCN, and an adversarial adaptive training mechanism. MIM captures semantic and structural relationships, LGG generates labels resembling ground truth, and LGD discriminates between generated and authentic labels for training.

We also enhance knowledge graph representation learning to map entities and relations into dense, low-dimensional vector spaces. Building on existing resources such as DBpedia \cite{auer2007dbpedia}, YAGC \cite{song2018fabrication}, and Freebase \cite{farber2018linked}, we extend representation learning to both one-hop and multi-hop neighbor subgraphs. In one-hop aggregation, both neighbor nodes and edges, as well as their influence on central node representations, are considered. Multi-hop subgraphs allow adaptive embedding of central nodes beyond the limitations of single-hop neighborhoods. These two levels of aggregation are fused into the Fat-RGCN module. 

We isolate and evaluate each module's contribution inside the enhanced knowledge representation framework using a series of ablation tests to confirm the efficacy of our suggested methodology.  To determine the value of integrating the two algorithmic improvements, we also analyze their combined performance.  To give thorough empirical proof of LabGraph's performance, extensive tests are conducted on the MIMIC-III benchmark dataset \cite{aldughayfiq2023capturing}.  According to Table~\ref{tab:experiment_results}, LabGraph outperforms current state-of-the-art techniques in every important evaluation criteria.

Main contributions of this study are summarized as follows:
\begin{itemize}
    \item Automatic EHR coding is reframed as a labeled graph generation problem, with LabGraph introduced as a unified multi-algorithm framework for ICD code prediction.
    \item MIM is proposed to explicitly model hierarchical parent-child relationships, sibling dependencies, and mutually exclusive code interactions.
    \item Four reinforcement learning–based training strategies are applied to enhance prediction of ICD codes.
    \item LGD delivers adversarial reward signals as intermediate supervision, guiding LabGraph to generate label sets that closely match the ground truth.
\end{itemize}

\section{METHOD AND THEORETICAL ANALYSIS}

\begin{figure}[ht]
\centerline{\includegraphics
[width=1\linewidth,keepaspectratio]
{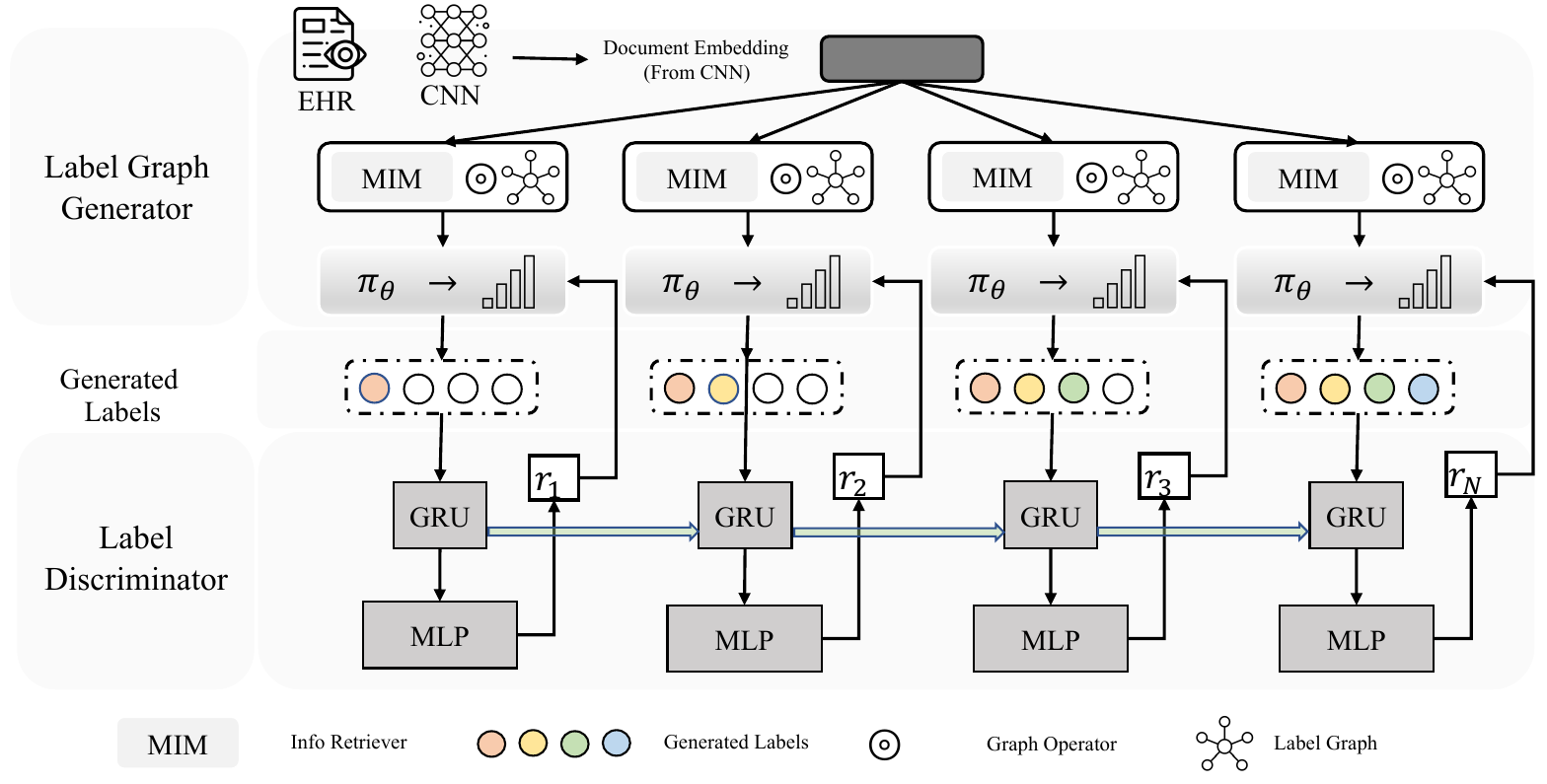}}
\caption{LabGraph Framework.}
\label{fig: architecture}
\end{figure}

Given a clinical text ${X} = \{x_1, \dots, x_N\}$, where $x_i$ denotes the $i$-th token, the objective is to predict a set of labels ${Y} = \{y^1, \dots, y^m\}$. Label graph generation starts from the root node and proceeds until a complete label cycle is formed. As illustrated in Figure~\ref{fig: architecture},{LabGraph} comprises two primary components: the {label graph generator} $G_{\theta}$ and the {label graph discriminator} $D_{\zeta}$.

\subsection{Meta-Parameter Learning}

{MCF:} Multi-Header Convolutional Filters capture patterns of varying durations. For filters with kernel sizes $k_1,...,k_n$:
\small
\begin{equation}
        F_{n}=f_{n}(X)=\bigwedge_{j=1}^{l} \tanh \left(W_{n}^{T} X^{j: j+k_{n}-1}\right).
\end{equation}

{MCB:} Multi-Residual Convolutional Blocks stack $p$ residual filters. Each block $c_{ni}$ applies three convolutions and aggregates as:
\small
\begin{equation}
        F_{ni}=\tanh \left(c_{ni_2}(I_1)+c_{ni_3}(I_1)\right).
\end{equation}

\subsection{Label Graph Generator $G_{\theta}$}
The label generation process is formulated as a Markov Decision Process (MDP), where states $\mathcal{S}$ correspond to predicted labels, actions $\mathcal{A}$ represent neighboring codes in the ICD hierarchy, and rewards $\mathcal{R}$ provide feedback for training via the REINFORCE algorithm \cite{williams1992simple}.
The expected return is:
\small
\begin{equation}
        \Bar{R}(\theta)=\Sigma_t \Sigma_{a_i\in \mathcal{A}} \pi(a_i|s_i,X;\theta)R(s_i,X,a_i).
\end{equation}
Policy $\pi(a_i|s_i,X;\theta)=\sigma(Ws_i+b_i)$ is updated via gradient ascent.

\subsection{Label Graph Discriminator $D_{\zeta}$}
Following \cite{DBLP:conf/sigir/WangRCRN0R20}, $D_{\zeta}$ evaluates generated paths ($c_1,...,c_i$). Using an LSTM encoder, the discrimination probability is:
\small
\begin{equation}
        h_i=\sigma(M_h(LSTM(h_{k-1},c_k)\oplus x)).
\end{equation}
Training minimizes cross-entropy with positive (ground truth) and negative (pseudo labels) samples.

\subsubsection{Multi-hop Model Integration (MHMI)}
We extend RGCN with attention and gating to exploit one-hop and multi-hop neighbors. For node $u$:
\small
\begin{equation}
    C_{u}= \left(1-D(C_{uj})\right)\cdot C_{uj}+D(C_{uj})\cdot C_{ui},
\end{equation}
where $D(\cdot)$ is a gate and $\beta_{O}$ denotes attention weights.

\subsection{Adversarial Adaptive Training (AAT)}
To enhance robustness, we apply adversarial perturbations \cite{jiang2019smart}. The objective is:
\small
\begin{equation}
        Loss=-\log p(y|r_{adv}+e;\zeta), \quad
        r_{adv}=\text{argmax}_{\Vert r \Vert < \epsilon}\log p(y|r+e;\zeta).
\end{equation}
We regularize with symmetrized KL-divergence to enforce smoothness.

\section{EXPERIMENTAL SETUP}
Experiments are conducted to address the research questions:

\subsubsection{Q1} How does LabGraph perform relative to current automated ICD coding in predicting ICD codes accurately?

\subsubsection{Q2} What training strategies can be employed for the label graph generation network to enhance its generalization, robustness, and overall effectiveness?

\subsection{Dataset}
MIMIC-III \cite{johnson2016mimic}: LabGraph’s validation uses the publicly available MIMIC-III dataset, which contains fewer than 50,000 records collected over 12 years since 2000. MIMIC-III complete is the full version, and MIMIC-III top 50 is the version with the top 50 codes.

Cora \cite{velivckovic2017graph}: Each node in the Cora graph is represented by a 1433-dimensional feature vector, with edges representing citation relationships between papers. Most tasks on Cora are node classification, though label sparsity exists in test graphs.

The FB15k-237 dataset \cite{schlichtkrull2018modeling} is a curated derivative of the Freebase knowledge repository \cite{li2025enhanced}, designed to provide a cleaner and more reliable benchmark for knowledge graph representation and reasoning tasks.
This paper is based on the FB15k-237 dataset that contains 14,541 nodes and 237 edge types, likewise use the same division as the baselines that are currently in use.

\subsection{Metric}
For consistency with earlier studies, model performance is evaluated using macro-averaged and micro-averaged F1 and AUC scores. The macro-averaged metrics are derived by first calculating the performance for each label independently and then taking the mean across all labels, providing an equal-weighted view of model behavior across both frequent and rare codes. Moreover, precision at K (P@K) is introduced to quantify how many of the top K predicted labels correspond to the true ICD codes, reflecting the model’s ranking accuracy. 

\subsection{Baselines}
To evaluate LabGraph, we compare it with several mainstream EHR coding models and hierarchical text classification methods:

Hierarchy-SVM \& Flat-SVMs \cite{nie2024towards}: Flat-SVMs constructs multiple independent binary classifiers for each ICD code using 10,000 TF-IDF features, treating codes in isolation and ignoring hierarchical structure. Hierarchy-SVM extends this approach by incorporating the ICD code hierarchy, enabling the model to exploit parent–child relationships and dependencies among codes for more structured predictions.

C-LSTM-Att \& C-MemNN \cite{zhang2023description, xi2025breaking}: C-LSTM-Att employs character-level LSTM models combined with attention mechanisms to better align clinical diagnosis descriptions with ICD codes, addressing mismatches caused by textual variability. C-MemNN iteratively condenses memory representations while retaining hierarchical information, effectively capturing both local semantic features and global code dependencies for improved prediction accuracy.

BI-GRU \cite{lee2020generating} \& HA-GRU \cite{dong2021explainable}: BI-GRU generates informative EHR embeddings for binary ICD classification; HA-GRU extends BI-GRU with hierarchical attention to improve disease categorization.

CAML \& DR-CAML \cite{li2021jlan}: CAML uses convolutional attention networks to learn ICD embeddings; DR-CAML adds label-wise normalization with regularization to enhance performance.

LAAT \& JointLAAT \cite{vu2020label}: LAAT applies label-specific attention on LSTM hidden states for ICD classification; JointLAAT adds hierarchical joint learning to improve efficiency and accuracy.

ISD~\cite{zhou2021automatic}, MSMN~\cite{yuan2022code}, \& FUSION~\cite{luo2021fusion}: ISD uses interactive shared representations to tackle long-tail issues; MSMN applies synonym matching and data augmentation to enhance embedding learning; FUSION addresses sparse diagnosis vocabularies via attention-based modeling of local and global features.

\subsection{Graph Generation Mechanism}
{ 
The ICD code generation process follows a hierarchical path from root to leaf nodes. Given input text X, at each step t:

1. State Encoding: Current state $s_t$ encodes the partial path $(c_1, ..., c_{t-1})$ and text features
2. Action Space Reduction: Valid actions $\mathcal{A}_t$ are restricted to children of $c_{t-1}$ in the ICD hierarchy
3. Policy Network: $\pi(a_t|s_t, X)$ selects next code based on text-code alignment scores
4. Termination: Generation stops when a leaf node is reached or a cycle forms

This formulation reduces the search space from O(|C|) to O(k), where k << |C| is the average branching factor, addressing the computational challenges of large label spaces.
}

\subsection{Computational Analysis}
{
Time complexity: $\mathcal{O}(L \cdot k \cdot d)$ where $L$ is path length, $k$ is branching factor, $d$ is embedding dimension. This compares favorably to flat classification $\mathcal{O}(|C| \cdot d)$ when $k \ll |C|$.

Space complexity: $\mathcal{O}(|V| + |E|)$ for storing the ICD graph, plus $\mathcal{O}(B \cdot L \cdot d)$ for batch processing, where $B$ is batch size.

Inference time on MIMIC-III: 0.23s per document (vs. 0.18s for CAML), showing practical efficiency despite increased model complexity.
}

\section{RESULT AND ANALYSIS}
ICD code distribution varies across levels: fourth-level codes account for 40.1\% of MIMIC-III Top50, while first- to third-level codes make up only 24.9\% of MIMIC-III Full. This suggests that navigating the ICD hierarchy from shallow to deep levels can reduce candidate search space, speed up inference, and improve LabGraph’s learning efficiency.

\subsection{Compared with Baselines (Q1)}
To address Q1, Table~\ref{tab:experiment_results}  presents LabGraph’s performance on MIMIC-III Full and Top50 datasets using core and personalized metrics. Key observations are:
First, LabGraph achieves the best results across all metrics, with low standard deviation, demonstrating both effectiveness and stability. Performance is better on MIMIC-III Full, indicating the model can capture deeper classification patterns and mitigate long-tail effects of rare diseases.
Second, compared to LabGraph, CAML, DR-CAML, LAAT, and JointLAAT show lower AUC and F1, reflecting poor coverage of rare codes and sparse ICD data. LabGraph reasons along the ICD hierarchy from root to leaf, reducing search space and improving inference efficiency.
Third, limitations of GRU-based models: BI-GRU and HA-GRU perform worse due to gradient vanishing in long EHR sequences, losing critical information. Disease-specific keywords are better captured by CNN-like models. LabGraph’s MHR-CNN module employs multi-headed CNNs with residual connections to obtain comprehensive EHR representations and mitigate gradient vanishing, improving classification accuracy.

\begin{table}[htbp]
    \caption{Experimental results on MIMIC-III Top-50 and Full datasets. LabGraph results are reported as mean $\pm$ standard deviation}
    \centering
    \resizebox{0.5\textwidth}{!}{
    \begin{tabular}{c|cc|cc|c|cc|cc|c}
    \toprule
    \multirow{3}{*}{\textbf{Model}} & \multicolumn{5}{c|}{\textbf{MIMIC-III Full}} & \multicolumn{5}{c}{\textbf{MIMIC-III Top50}} \\\cline{2-11}
     & \multicolumn{2}{c|}{\textbf{\textit{AUC}}} & \multicolumn{2}{c|}{\textbf{\textit{F1}}} & \multirow{2}{*}{P@8} & \multicolumn{2}{c|}{\textbf{\textit{AUC}}} & \multicolumn{2}{c|}{\textbf{\textit{F1}}} & \multirow{2}{*}{\textbf{\textit{P@5}}} \\\cline{2-5} \cline{7-10}
     & \textbf{\textit{Macro}} & \textbf{\textit{Micro}} & \textbf{\textit{Macro}} & \textbf{\textit{Micro}} & & \textbf{\textit{Macro}} & \textbf{\textit{Micro}} & \textbf{\textit{Macro}} & \textbf{\textit{Micro}} & \\
     \hline
    Hierarchy-SVM & 0.456 & 0.438 & 0.009 & 0.001 & 0.202 & 0.376 & 0.368 & 0.041 & 0.079 & 0.144 \\
    Flat-SVMs & 0.482 & 0.467 & 0.011 & 0.002 & 0.242 & 0.439 & 0.401 & 0.048 & 0.093 & 0.179 \\ 
    \hline
    C-MemNN & 0.833 & 0.913 & 0.082 & 0.514 & 0.695 & 0.824 & 0.896 & 0.509 & 0.588 & 0.596 \\
    C-LSTM-Att & 0.831 & 0.908 & 0.079 & 0.511 & 0.687 & 0.816 & 0.892 & 0.501 & 0.575 & 0.574 \\
    \hline
    BI-GRU & 0.500 & 0.547 & 0.002 & 0.140 & 0.317 & 0.501 & 0.594 & 0.035 & 0.268 & 0.228 \\
    HA-GRU & 0.501 & 0.509 & 0.017 & 0.004 & 0.296 & 0.500 & 0.436 & 0.072 & 0.124 & 0.205 \\
    \hline
    CAML & 0.895 & 0.959 & 0.088 & 0.539 & 0.709 & 0.875 & 0.909 & 0.532 & 0.614 & 0.609 \\
    DR-CAML & 0.897 & 0.961 & 0.086 & 0.529 & 0.609 & 0.884 & 0.916 & 0.576 &0.633 & 0.618 \\
    \hline
    LAAT & 0.919 & 0.963 & 0.099 & 0.575 & 0.738 & 0.925 & 0.946 & 0.666 & 0.715 & 0.675 \\
    JointLAAT & 0.941 & 0.965 & 0.107 & 0.577 & 0.735 & 0.925 & 0.946 & 0.661 & 0.716 & 0.671 \\
    \hline
    ISD & 0.938 & \uline{0.967} & \uline{0.119} & 0.559 & 0.745 & \uline{0.935} & \uline{0.949} & 0.679 & 0.717 & \uline{0.682} \\
    MSMN & \uline{0.943} & 0.965 & 0.103 & 0.584 & \uline{0.752} & 0.928 & 0.947 & \uline{0.683} & \uline{0.725} & 0.680 \\
    FUSION & 0.915 & 0.964 & 0.088 & \uline{0.636} & 0.736 & 0.909 & 0.933 & 0.619 & 0.674 & 0.647 \\
    \hline
    \multirow{3}{*}{LabGraph} & \textbf{0.989} & \textbf{0.998} & \textbf{0.134} & \textbf{0.789} & \textbf{0.798} & \textbf{0.981} & \textbf{0.989} & \textbf{0.754} & \textbf{0.787} & \textbf{0.763} \\
    
     & \textcolor[RGB]{51,153,102}{(+4.88\%)} & \textcolor[RGB]{51,153,102}{(+3.21\%)} & \textcolor[RGB]{51,153,102}{(+12.61\%)} & \textcolor[RGB]{51,153,102}{(+19.39\%)} & \textcolor[RGB]{51,153,102}{(+6.12\%)} & \textcolor[RGB]{51,153,102}{(+4.91\%)} & \textcolor[RGB]{51,153,102}{(+4.21\%)} & \textcolor[RGB]{51,153,102}{(+7.10\%)} & \textcolor[RGB]{51,153,102}{(+8.55\%)} & \textcolor[RGB]{51,153,102}{(+11.88\%)} \\
     
     & $\pm$ 0.002 & $\pm$ 0.001 & $\pm$ 0.001 & $\pm$ 0.002 & $\pm$ 0.001 & $\pm$ 0.001 & $\pm$ 0.002 & $\pm$ 0.001 & $\pm$ 0.002 & $\pm$ 0.001 \\
    \bottomrule
    \end{tabular}}
    \label{tab:experiment_results}
\end{table}
     
\begin{table}[htbp]
    \caption{Ablation study results on the MIMIC-III Top-50 and Full datasets. Standard deviations for LabGraph are omitted as they are consistent with those in the previous table.}
    \centering
    \resizebox{0.5\textwidth}{!}
    {
    \begin{tabular}{c|cc|cc|c|cc|cc|c}
    \hline
    \multirow{3}{*}{\textbf{Model}} & \multicolumn{5}{c|}{\textbf{MIMIC-III Full}} & \multicolumn{5}{c}{\textbf{MIMIC-III Top50}} \\\cline{2-11}
     & \multicolumn{2}{c|}{\textbf{\textit{AUC}}} & \multicolumn{2}{c|}{\textbf{\textit{F1}}} & \multirow{2}{*}{P@8} & \multicolumn{2}{c|}{\textbf{\textit{AUC}}} & \multicolumn{2}{c|}{\textbf{\textit{F1}}} & \multirow{2}{*}{\textbf{\textit{P@5}}} \\\cline{2-5} \cline{7-10}
     & \textbf{\textit{Macro}} & \textbf{\textit{Micro}} & \textbf{\textit{Macro}} & \textbf{\textit{Micro}} & & \textbf{\textit{Macro}} & \textbf{\textit{Micro}} & \textbf{\textit{Macro}} & \textbf{\textit{Micro}} & \\
    \hline
    LabGraph & \textbf{0.983} & \textbf{0.998} & \textbf{0.134} & \textbf{0.622} & \textbf{0.798} & \textbf{0.981} & \textbf{0.989} & \textbf{0.754} & \textbf{0.787} & \textbf{0.763} \\
    \hline
    No ARCL & 0.834 & 0.867 & 0.098 & 0.509 & 0.645 & 0.813 & 0.852 & 0.594 & 0.619 & 0.521 \\
    \hline
    No MHR-CNN & 0.862 & 0.901 & 0.099 & 0.515 & 0.659 & 0.833 & 0.889 & 0.637 & 0.629 & 0.573 \\
    \hline
    \end{tabular}}
    \label{tab:ablation_experiment_results}
\end{table}

\subsection{LabGraphAblation (Q2)}
A series of ablation experiments were conducted to assess the contribution of each module within LabGraph, with the corresponding results presented in Table~\ref{tab:ablation_experiment_results}. Four model variants were examined:
(1) No ARCL, which excludes adversarial reinforcement causal learning in both the LGG and LGD components, thereby training the network solely on teacher-forced and ground-truth paths;
(2) No MIM, which omits the Message Integration Module, preventing the transmission of hierarchical relationships such as parent–child and sibling dependencies;
(3) No MHR-CNN, which substitutes the MHR-CNN with a conventional CNN, leading to less precise feature embeddings; and
(4) No AAT, which removes adversarial adaptive training, eliminating the embedding layer’s perturbation-based robustness enhancement. 

As shown in Table~\ref{tab:ablation_experiment_results}, 1) No ARCL drastically degrades performance. On MIMIC-III Full, macro/micro AUC drop by 15.16\%/13.13\%, macro/micro F1 by 26.87\%/18.17\%, and P@8 by 19.17\%; similar trends occur on Top50. This indicates ARCL is crucial for generating ICD code paths and calculating rewards. 2) No MIM also reduces performance. On Top50, macro/micro AUC drop by 9.07\%/5.97\%, macro/micro F1 by 10.61\%/8.77\%, and P@5 by 13.70\%, showing that MIM captures interactions between EHR text and ICD codes. 3) No MHR-CNN causes average declines of 11.02\% in AUC and 21.66\%/17.42\% in macro/micro F1 on Full, demonstrating that multi-head CNN with residual connections better captures EHR features and mitigates gradient vanishing. 4) No AAT decreases performance as well; on Top50, macro/micro AUC drop by 5.40\%/5.16\%, macro/micro F1 by 7.29\%/7.88\%, and P@5 by 11.66\%, confirming that AAT enhances embedding representation and stabilizes training.

\subsubsection{Component-wise Performance Analysis}
To better understand the contribution of each component, we conducted progressive ablation studies. 
Starting with a baseline CNN encoder, we observe the following incremental gains:

  \begin{itemize}
    \item {MHR-CNN}: +12.3\% (multi-scale feature extraction)
    \item {MIM}: +8.7\% (hierarchical relationship modeling)
    \item {ARCL}: +15.2\% (adversarial training)
    \item {AAT}: +6.1\% (robustness enhancement)
\end{itemize}

Overall, each module contributes, with ARCL providing the largest performance gain.

\subsubsection{Sensitivity Analysis}
{We analyzed c's sensitivity to key hyperparameters. Varying the adversarial perturbation radius $\epsilon$ from 0.01 to 1.0 shows optimal performance at $\epsilon=0.1$, with degradation beyond 0.5 due to excessive noise. The number of residual blocks in MHR-CNN peaks at 3 blocks, with diminishing returns thereafter.
}

\subsection{Qualitative Analysis}
\subsubsection{Case Study}
Consider a discharge summary mentioning ``patient admitted with chest pain, diagnosed with acute myocardial infarction, developed heart failure during stay.''

Traditional models often miss the causal relationship, predicting only primary codes. LabGraph's generation path:
\begin{enumerate}
    \item Root $\rightarrow$ Circulatory System (390-459)
    \item $\rightarrow$ Ischemic Heart Disease (410-414) 
    \item $\rightarrow$ Acute Myocardial Infarction (410)
    \item $\rightarrow$ Heart Failure (428) [via sibling traversal]
\end{enumerate}
This demonstrates how graph generation captures both hierarchical and lateral relationships.

\subsubsection{Error Analysis}
Common failure modes include:
\begin{itemize}
    \item Rare disease codes ($<$ 10 occurrences): 42\% error rate
    \item Ambiguous terminology: 31\% of errors
    \item Missing clinical context: 27\% of errors
\end{itemize}

\section{CONCLUSION AND FUTURE WORK}

In this work, the task of encoding and classifying electronic health records (EHRs) is reformulated as the generation of adversarial hierarchical label graphs, capturing both the complex textual features of clinical notes and the structured dependencies of ICD codes. The proposed framework, LabGraph, leverages adversarial transfer learning and integrates MHR-CNN and Fat-RGCN modules to extract intricate semantic patterns from medical text and hierarchical label structures. A Message Integration Module (MIM) is further incorporated to explicitly model relationships among ICD codes, including parent–child and sibling dependencies, enhancing the model’s structural awareness. Extensive experiments on the MIMIC-III dataset demonstrate that LabGraph consistently surpasses existing competitive approaches across multiple evaluation metrics, establishing a new benchmark for automated ICD coding.

While LabGraph achieves superior results, several limitations remain. First, scalability to the full ICD-10 ontology with over 70,000 codes remains challenging. Second, the interpretability of neural components is limited, which may hinder clinical adoption. Third, performance still depends heavily on large-scale annotated data, and degrades significantly with limited training samples. Finally, cross-institution transfer is difficult, as variations in coding practices often require extensive retraining.  

Future work will explore the combination of pre-trained medical language models such as BioBERT with the graph pre-training methods to improve generalization. We also plan to investigate few-shot learning methods for rare disease codes, develop automated hyperparameter tuning strategies, and design explainable AI techniques to enhance clinical interpretability. In addition, optimizing loss functions and incorporating prior medical knowledge hold promise for building more robust and generalizable ICD coding systems.


\bibliographystyle{ieeetr}   
\normalem

\bibliography{sample-base2}   

\end{document}